# A robust approach for tree segmentation in deciduous forests using small-footprint airborne LiDAR data

Hamid Hamraz[a*], Marco A. Contreras[b], Jun Zhang[a]

a: Department of Computer Science, b: Department of Forestry

University of Kentucky, Lexington, KY 40506, USA

hhamraz@cs.uky.edu, marco.contreras@uky.edu, jzhang@cs.uky.edu

* Corresponding Author: hhamraz@cs.uky.edu +1 (859) 489 1261

**Abstract.**

This paper presents a non-parametric approach for segmenting trees from airborne LiDAR data in deciduous forests. Based on the LiDAR point cloud, the approach collects crown information such as steepness and height on-the-fly to delineate crown boundaries, and most importantly, does not require a priori assumptions of crown shape and size. The approach segments trees iteratively starting from the tallest within a given area to the smallest until all trees have been segmented. To evaluate its performance, the approach was applied to the University of Kentucky Robinson Forest, a deciduous closed-canopy forest with complex terrain and vegetation conditions. The approach identified 94% of dominant and co-dominant trees with a false detection rate of 13%. About 62% of intermediate, overtopped, and dead trees were also detected with a false detection rate of 15%. The overall segmentation accuracy was 77%. Correlations of the segmentation scores of the proposed approach with local terrain and stand metrics was not significant, which is likely an indication of the robustness of the approach as results are not sensitive to the differences in terrain and stand structures.

# 1 Introduction

Traditionally, decision making in forest management has been based on stand attribute information collected using stand examinations and topographic surveys. Because field-based inventory data are expensive and labor-intensive to acquire, sampling intensity of field inventory is typically limited, providing rough estimates of stand attributes while ignoring large variability in terrain and vegetation within stands (Shiver and Borders, 1996). In the last two decades, light detection and ranging (LiDAR) technology has brought drastic changes to forest data acquisition and management by providing inventory data at unprecedented spatial and temporal resolutions (Ackermann, 1999; Maltamo et al., 2014; Shao and Reynolds, 2006; Wehr and Lohr, 1999). However, in order to obtain accurate tree level attributes such as crown width and tree height as well as derivative estimates (i.e., diameter at breast height (DBH), volume, and biomass), accurate and automated tree segmentation approaches are required (Schardt et al., 2002).

Numerous methods have been developed to segment individual trees from LiDAR data. Earlier methods used pre-processed data in the form of raster digital surface models (DSMs) or canopy height models (CHMs) and more recent methods directly used the LiDAR point clouds (Hyyppä et al., 2001; Persson et al., 2002; Wang et al., 2008). Regardless of input, existing tree segmentation methods can be categorized as parametric and non-parametric. In general, parametric methods fit 3D shape models (Holmgren et al., 2010) or perform multi-stage filtering (Falkowski et al., 2006; Pitkänen et al., 2004; Wolf and Heipke, 2007), where the filtering kernel functions or the shape models are assumed to adequately estimate the geometric shapes representing the tree crowns. The parameters defining these functions/models are set manually based on typical tree crown shapes and sizes obtained previously from field sampling. A recent multi-stage filtering method (Jing et al., 2012) applies a series of morphological opening operations (Serra, 1986; Soille, 2003) to determine the dominant sizes of the tree crowns, allowing the parameters of the filter kernel to be set automatically for each stage. Although this method avoid manually setting parameters, selecting appropriate kernel function and combining the result of different stages is non-trivial, especially in natural forests with highly variable crown shapes and sizes.

Non-parametric methods identify local maxima (LMXs), assumed as the tree apexes, or use local minima (LMs) to find tree crown boundaries. LMX-based methods search for tree apexes within a neighborhood window and then perform a variety of region-growing or clustering routines to delineate tree crowns (Alizadeh Khameneh, 2013; Hyyppä et al., 2001; Morsdorf et al., 2004; Persson et al., 2002; Véga and Durrieu, 2011). Determining the size of the neighborhood window to search for the tree apexes is non-trivial and can easily result in missing apexes or identifying false trees. A widely used approach is to adaptively size the window based on the tree height using site-specific regression models (Chen et al., 2006; Popescu and Wynne, 2004; Popescu et al., 2002). However, this approach works well only when trees are homogeneously shaped where an accurate crown width model based on trees height only can be created (Pitkänen et al., 2004). More recent methods perform multi-stage non-parametric segmentation (Véga and Durrieu, 2011; Véga et al., 2014) where a variety of window sizes are used to create segmentation maps at different scales. The results of the different stages are then incorporated according to a scoring system based on different properties of an ideal crown shape. Li et al. (2012) resolved the problem of correctly identifying LMX by assuming the highest non-clustered LiDAR point represented the apex of the tallest tree, however, the clustering method was also considering only the vegetation height.

LM-based methods typically use watershed segmentation routines (Beucher and Lantuéjoul, 1979; Vincent and Soille, 1991) to detect crown boundaries and perform subsequent valley following routines to find the area representing individual tree crowns (Gougeon, 1995; Leckie et al., 2003). In general, watershed segmentation is prone to under/over-segmentation due to differences in tree heights and natural variability of vegetation within tree crowns. To overcome this problem, studies use marker-controlled watershed segmentation routines (Dougherty et al., 2003), where the basic idea is to mark the trees and guide the watershed procedure to only delineate those marked trees. Marking manually (Chen et al., 2006) is impractical for large-scale data. Automated approaches have generally marked the tree apexes by performing morphological image analysis (Kwak et al., 2007). Similar to LMX-based methods, these automated approaches are prone to missing tree apexes or identifying false ones, especially when trees are not homogeneously shaped and sized. Several methods have used a combination of apex identification (LMX-based) and watershed segmentations (LM-based) to perform crown delineation and thus improve tree detection rates (Chen et al., 2006; Hu et al., 2014; Jing et al., 2012; Kwak et al., 2007).

Existing tree segmentation methods have mostly focused on conifer forests or forests with relatively open canopy, where assumptions about size and shape of tree crowns and/or spacing among trees are made (Heurich, 2008; Kaartinen et al., 2012). These assumptions make the methods forest type specific and not easily applicable to forests with different conditions (Vauhkonen et al., 2011). Deciduous forests present significantly more complex vegetation conditions due to large variation in tree shapes and sizes, larger number of species and denser canopy, where individual trees are much harder to detect (Heurich, 2008; Jing et al., 2012; Koch et al., 2006; Véga et al., 2014). Studies report that performance of previous methods varies drastically from 50% to over 90% of tree detection accuracy depending on the forest conditions and types, species distribution, and stand structure (Kaartinen et al., 2012; Vauhkonen et al., 2011). These results suggest that there is no universally superior method and that these methods are custom designed for specific vegetation conditions, which evidence the need to develop general approaches that can be applied to multiple forest types while ensuring robust tree detection results.

In this paper, we present a robust and novel approach for segmenting trees within small-footprint LiDAR data in deciduous forests with complex terrain and vegetation conditions. The method is non-parametric and delineates individual tree crowns based on only the local information, crown shape and height of the vegetation, and does not require a priori knowledge of either stand structure or typical tree attributes. A major improvement of our approach, compared with existing approaches, is the dynamic capture of local information about crown shape and its use to enhance crown delineation.

## 2 Tree segmentation approach

The main inputs of the tree segmentation approach are the LiDAR point cloud and the LiDAR-derived DEM. Independent of the point density, LiDAR point clouds have variable, small-scale point spacing resulting from scan patterns (e.g., zig-zag) and flight line overlap. Thus, a pre-processing routine is applied to homogenize point spacing. This routine creates a grid with resolution equal to the average nominal post spacing (NPS) and filters the LiDAR point cloud by selecting the highest elevation LiDAR point within each grid cell, hereafter called LiDAR surface points (LSPs). Using the LiDAR-derived DEM, heights above ground are calculated for all LSPs. Those LSPs below a minimum height, set here as 5 meter, are removed from further analysis. Based on the vegetation structure (stem density and variability in tree heights), this creates several gaps with no vegetation in the remaining LSP dataset, which is utilized later in the analysis. The last pre-processing step smooths LSPs to reduce small variation in vegetation elevation within tree crowns while maintaining important vegetation patterns. A Gaussian smoothing filter with standard deviation equal to the NPS and a radius of 3×NPS was used.

After the pre-processing steps, the tree segmentation approach consists of the following routines: 1) locate the global maximum elevation (GMX) amongst LSPs, which is assumed to represent the apex of the tallest tree within a given area, 2) generate vertical profiles originating from the GMX location and expanding outwards, 3) identify the individual LSP along the profile that likely represents the crown boundary using between-tree gap identification and LM identification for each profile, 4) create a convex hull of boundary points, which delineates the tree crown, and 5) cluster all LSPs encompassed within the convex hull and assign them as the current tallest tree crown. This process is applied iteratively until all LSPs have been clustered into tree crowns. Clusters representing crowns with diameter below a minimum detectable crown width (MDCW), set here as 1.5 m, are considered noise. Figure 1 shows the flowchart of the tree segmentation approach and Figure 2 shows an example of the application of the five routines within the approach.

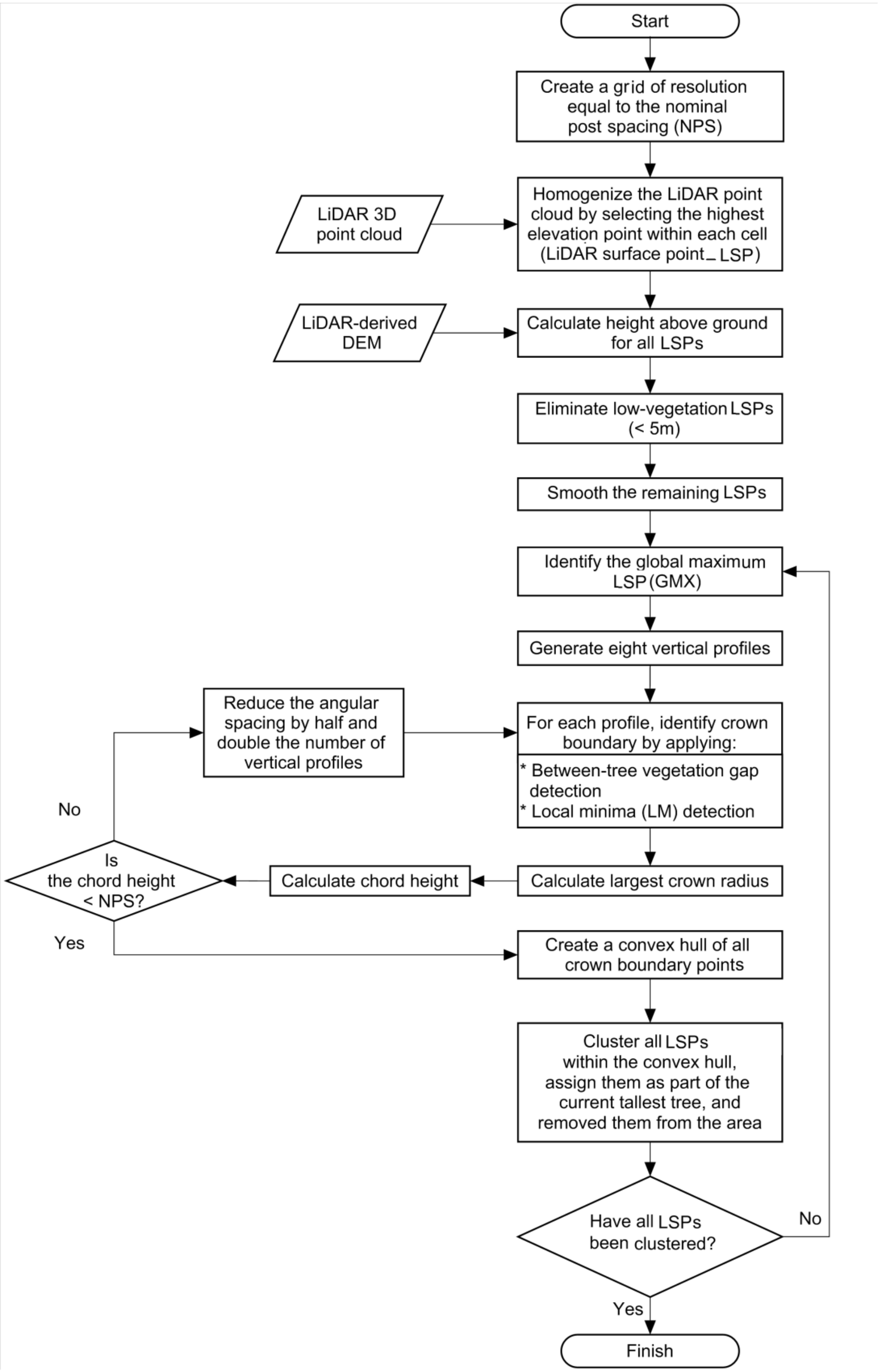

Figure 1. Flowchart of the tree segmentation approach used to identify tree locations and delineate tree crowns.

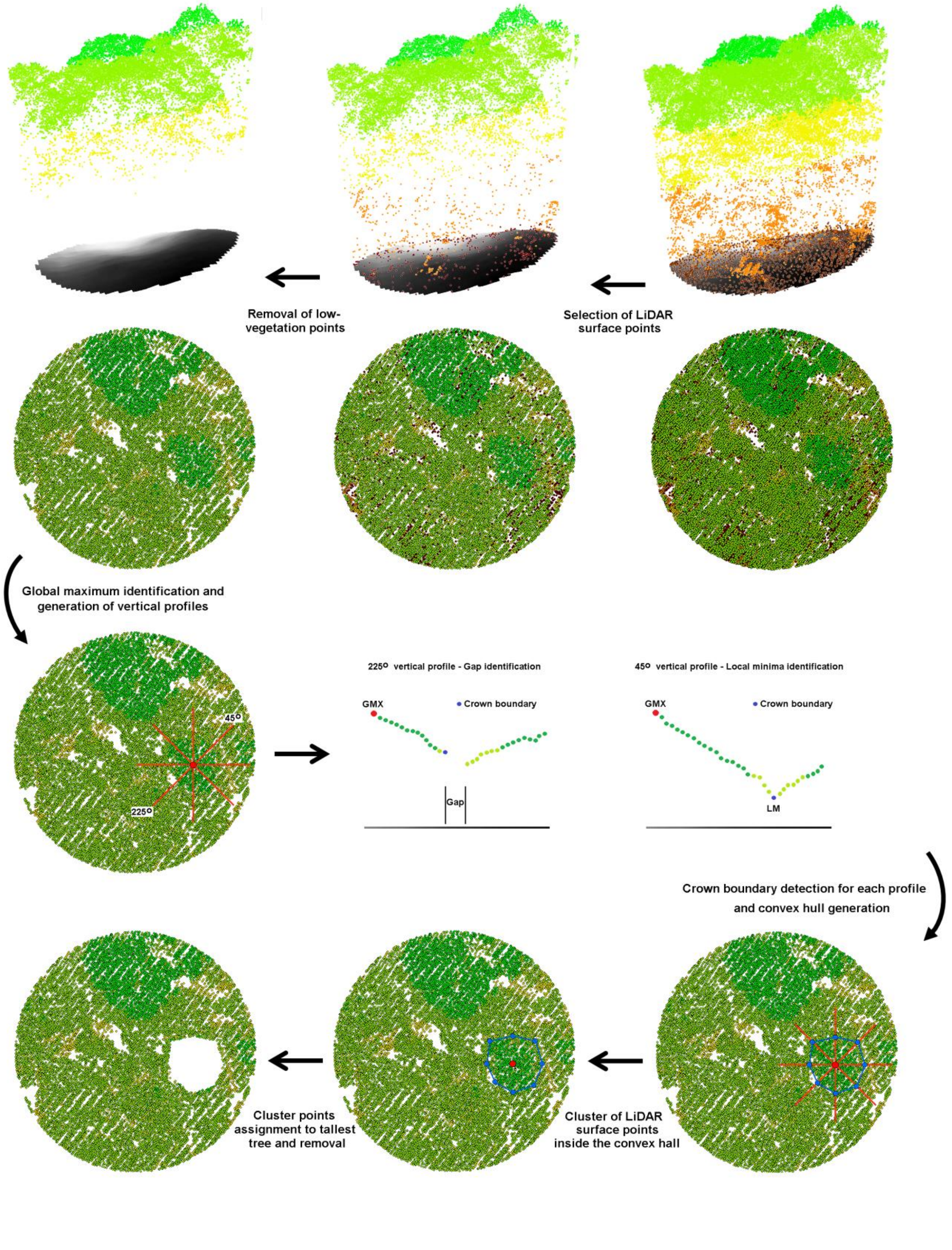

Figure 2. Illustration of the preprocessing steps and the five routines within the tree segmentation approach.

The most critical and non-trivial routines of the tree segmentation approach are the generation of an appropriate number of profiles and the identification of crown boundary points to accurately delineate tree crowns. The procedures developed for these two routines form the basis for this novel tree segmentation approach.

## 2.1 Profile generation

After identifying the GMX within a given area, vertical profiles originating from it and expanding a maximum horizontal distance, set here to 15.24 m (50 feet), are generated. The number of profiles required to smoothly represent tree crowns is determined dynamically based on LiDAR-detected crown radii. The procedure starts with eight uniformly spaced profiles (every 45°). After the crown boundary and thus radius is determined for each profile (explained below), the maximum crown radius ($r$) is used to determine the chord height ($x$) between two maximum crown radius profiles separated by the angular spacing ($\varphi$) (Figure 3) as follows:

$$x = r(1 - \cos(\varphi/2)) \quad [1]$$

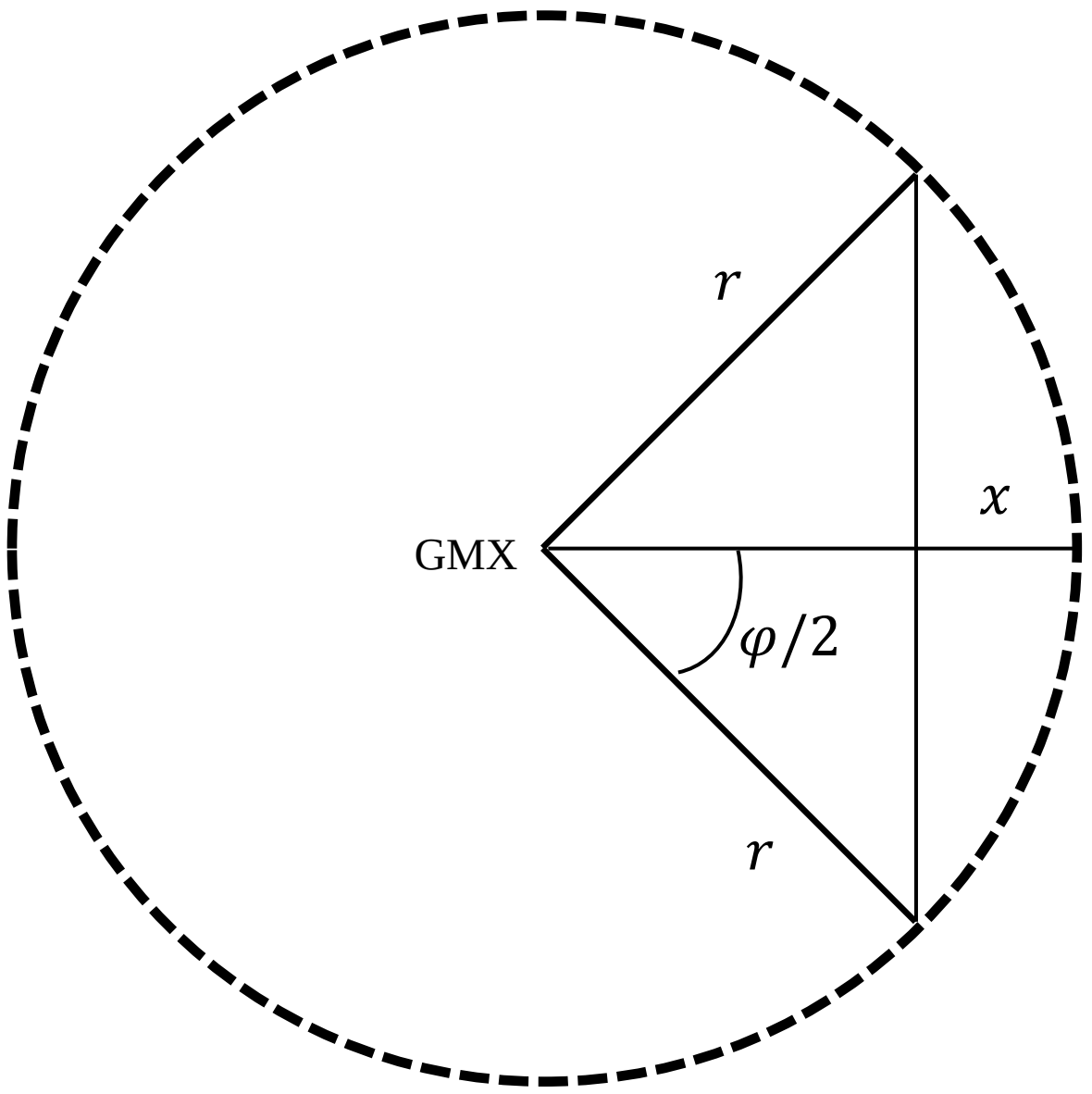

Figure 3. Diagram illustrating the calculation of the chord height ($x$) formed by two profiles of maximum crown radius ($r$) separated by the angular spacing ($\varphi$).

If the chord height is larger than NPS, the angular spacing is reduced by half and the number of profiles is doubled. The new chord height is calculated again based on the updated maximum crown radius and the new profile angular spacing. Doubling the number of profiles continues iteratively until the chord height is smaller than NPS. By using the maximum LiDAR-detected crown radius, the procedure ensures a sufficiently large number of profiles and thus a smooth delineation of the tree crown.

The width of each profile was set to 2×NPS to ensure a sufficient number of LSPs representing vegetation characteristics. Profiles are then analyzed vertically in two dimensions using horizontal distance from the GMX and the elevation associated with each LSP.

## 2.2 Crown boundary identification

After generating a vertical profile and identifying all LSPs along it, two sub-routines are applied to identify the crown boundary. The first sub-routine identifies inter-tree crown gaps via statistical analysis of the distribution of horizontal distances between consecutive points along the profile. Thereafter, the second sub-routine inspects LM points as potential crown boundaries based on the median slope of points within two windows expanding both directions from each LM location.

### 2.2.1 Identifying inter-tree gaps

This sub-routine identifies relatively large horizontal distance between any two consecutive LSPs along the profile, which are assumed to represent vegetation gaps between adjacent tree crowns. For this purpose, the Tukey statistical outlier detection method is used (McGill et al., 1978). Horizontal distances between consecutive points along the profile follow a Poisson distribution. When transformed to their square root, the resulting distribution can reasonably approximate a normal distribution (Thacker and Bromiley, 2001), which is more appropriate for the Tukey method. To be conservative, square root distance values larger than six times the interquartile range from the third quartile is considered inter-tree gaps. After gaps are identified, only the LSPs between the GMX and the first gap remain for further analysis and the LSPs located beyond the first gap are removed from the profile. At this point, it is assumed that the remaining LSPs represent either the crown of the current tallest tree or the crowns of multiple adjacent trees, growing close together with overlapping crowns.

### 2.2.2 Identifying LM points as crown boundary

Starting from the GMX, this sub-routine identifies LM points defined as those with elevations lower than their two adjacent neighbors. Once an LM point is found, the sub-routine determines whether it represents the crown boundary or natural variation of vegetation height within the crown. For this purpose, two windows expanding on both sides of the LM are created. The left window considers all LSPs from the GMX to the LM. The size of the right window is estimated based on the: i) steepness of consecutive points within a distance equal to MDCW on the right of the LM, and ii) crown radii of two hypothetical trees of equal height crowns of which represented by two distinctly different shapes (a sphere and a narrow cone).

The steepness of LSPs on the right of the LM ($S_{right}$) is calculated as the median (in degrees) of absolute slopes between consecutive points ($i, i+1$) within MDCW meters from the LM ($w_{MDCW}$):

$$S_{right} = \tan^{-1}\left(median\left[\left|slope_{i,\ i+1}\right|; i\ , i+1 \epsilon w_{MDCW}\right]\right) \qquad [2]$$

If the LM is in fact the crown boundary, the LSPs within $w_{MDCW}$ partially represent the crown of an overlapping and shorter tree with a steepness that is approximated by $S_{right}$. The value of $S_{right}$ should range between the steepness of a sphere-shaped crown and the steepness of a narrow cone-shaped crown (two ends of the spectrum). As the height of the adjacent tree ($h_{ad}$) is between the heights of the GMX and the LM point, its height is reasonably approximated by the average of the GMX and the LM heights.

The steepness of a narrow cone-shaped crown can be expressed as 90°-ϵ, whereϵ (set here as 5°) indicates a small deviation from vertical. The cone-shaped crown radius ($cr_c$) can then be calculated as follows:

$$cr_c = \frac{h_{ad} \times CL_c}{tan(90° - \varepsilon)} \times O_c \qquad [3]$$

where, $CL_c$ is the crown ratio, and $O_c$ indicates the crown radius reduction due to the overlap assuming the narrow cone-shaped tree is situated in a dense stand.

On the other hand, the slope of a sphere-shaped crown ranges from 0° to 90° with the steepness (expected value) of 32.7° (see Appendix A). Its crown radius ($cr_s$) can be calculated as follows:

$$cr_s = \frac{h_{ad} \times CL_s}{2} \times O_s \qquad [4]$$

where, $CL_s$ and $O_s$ indicate the crown ratio and the crown radius reduction due to the overlap within a dense stand for the sphere-shaped tree.

Then, the size of the right window ($w_{rd}$) is calculated by interpolating $cr_c$ and $cr_s$ with respect to $S_{right}$ (which should be between 32.7° and 90°-ϵ):

$$w_{rd} = cr_c\left(1 - \frac{(90-\varepsilon) - S_{right}}{(90-\varepsilon) - 32.7}\right) + cr_s\left(\frac{(90-\varepsilon) - S_{right}}{(90-\varepsilon) - 32.7}\right) \qquad [5]$$

Lastly, after determining both window sizes on either side of the LM, the median of slopes between consecutive LSPs of each window is calculated. If the median slope of the left-side window is negative (downwards from the apex to the crown boundary) and the median slope of the right-side window is positive (upwards from the crown boundary toward the apex of the adjacent tree crown), then the LM is considered a boundary point. Otherwise, the current LM is considered to represent natural variation of vegetation height within the current tallest tree crown and the next LM farther from the GMX along the profile is evaluated. If none of the LMs found meet the crown boundary criterion then the last LSP is considered as the crown boundary.

Crown ratio is highly variable among individual trees and species dependent with values typically varying between 0.4 and 0.8 (Randolph, 2010). The crown ratio of a narrow cone-shaped tree tends to be larger than that of a sphere-shaped one (Kim et al., 2009). So, for the purpose of illustrating the application of our approach, we used 0.8 and 0.7 for $\mathrm{CL_c}$ and$\mathrm{CL_s}$, respectively. Similarly, crown radius reduction due to overlap is highly variable with a value of about 0.5 for a really dense stand. The radius of a narrow cone-shaped tree tends to be reduced less than of a sphere-shaped tree because the crown of a narrow cone-shaped tree is quite compact from the sides. So, we used two thirds for $\mathrm{O_c}$ and one third for$\mathrm{O_s}$. Although the constant values set here can affect the final size determined for the right window (Equation 5), the sign of the median slope would be the same as long as the size is within a reasonable range. Still possible in practice, an excessively narrow window might result in erroneously flipping the sign of the median slope and an LM representing natural vegetation height within the crown to be misidentified as the crown boundary and vice versa. However, when considering the multiple profiles generated for each GMX, the effect of a single window size on the ability to delineate tree crown is minimal.

Both sub-routines, to identify inter-tree gaps and crown boundaries respectively, are completely based on the 3D positions of LSPs along a profile. This avoids prior assumptions of tree crown shapes and dimensions, which makes the approach a robust method that can be applied to different vegetation types.

# 3 Approach Application

## 3.1 Study site and field data

The tree segmentation approach application was conducted at the University of Kentucky's Robinson Forest (RF, Lat. 37.4611, Long. -83.1555) located in the rugged eastern section of the Cumberland Plateau region of southeastern Kentucky in Breathitt, Perry, and Knott counties (Figure 4). The terrain across RF is characterized by a branching drainage pattern, creating narrow ridges with sandstone and siltstone rock formations, curving valleys and benched slopes. The slopes are dissected with many intermittent streams (Carpenter and Rumsey, 1976) and are moderately steep ranging from 10 to over 100% facings predominately northwest and south east, and elevations ranging from 252 to 503 meters above sea level. Vegetation is composed of a diverse contiguous mixed mesophytic forest made up of approximately 80 tree species with northern red oak (*Quercus rubra*), white oak (*Quercus alba*), yellow-poplar (*Liriodendron tulipifera*), American beech (*Fagus grandifolia*), eastern hemlock (*Tsuga canadensis*) and sugar maple (*Acer saccharum*) as dominant and codominant species. Understory species include eastern redbud (*Cercis canadensis*), flowering dogwood (*Cornus florida*), spicebush (*Lindera benzoin*), pawpaw (*Asimina triloba*), umbrella magnolia *(Magnolia tripetala*), and bigleaf magnolia (*Magnolia macrophylla*) (Carpenter and Rumsey, 1976; Overstreet, 1984). Average canopy cover across Robinson Forest is about 93% with small opening scattered throughout. Most areas exceed 97% canopy cover and recently harvested areas have an average cover as low as 63%. After being extensively logged in the 1920's, Robinson Forest is considered second growth forest ranging from 80-100 years old, and is now protected from commercial logging and mining activities, typical of the area (Department of Forestry, 2007).

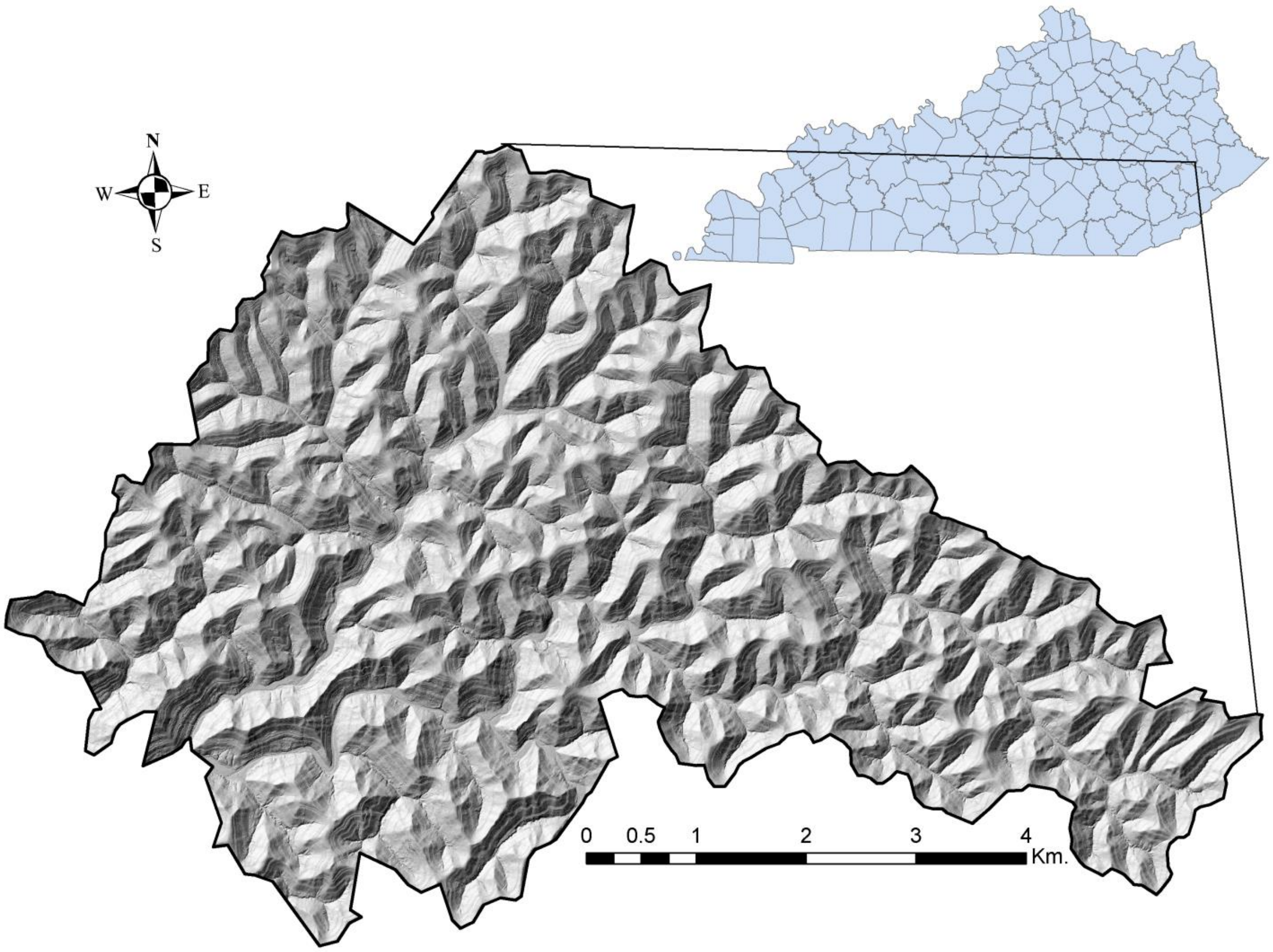

Figure 4. Terrain relief map of the University of Kentucky Robinson Forest and its general location within Kentucky, USA.

Within the Clemons Fork watershed (which covers an area of about 1,500 ha), 1.2×1.2 meter plywood boards, painted white to increase reflectance were installed prior the acquisition of LiDAR data at 103 existing circular permanent plots each with an area of 0.1 ac (~0.04 ha). Boards were installed, leveled with their centers placed at the exact location of the plot rebar markers, with the purpose of correctly geo-referencing the exact location of plot centers. Within each plot, DBH (cm), tree height (m), species, crown class (dominant, co-dominant, intermediate, overtopped), tree status (live, dead), and stem class (single, multiple) were recorded for all trees with DBH > than 12.5 cm. In addition, horizontal distance and azimuth from plot center to the face of each tree at breast height were collected to create a stem map. Site variables including slope, aspect, and slope position were also recorded for each plot.

## 3.2 LiDAR data

We combined two LiDAR datasets covering the study area, collected with the same LiDAR system by the same vendor. One dataset was low density (~1.5 pt/m$^2$) collected in the spring of 2013 during leaf-off season for the purpose of acquiring terrain information, as a part of a state-wide elevation data acquiring program from the Kentucky Division of Geographic Information. The second dataset was high density (~25 pt/m$^2$) acquired in the summer of 2013 during leaf-on season for the purpose of collecting detailed vegetation information and ordered by the University of Kentucky Department of Forestry. The parameters of the LiDAR system and flight for both datasets are presented in Table 1. The vendor processed both raw LiDAR datasets using the TerraScan software (Terrasolid Ltd., 2012) to classify LiDAR points into ground and non-ground points. The LASTools (Isenburg, 2011) extension in ArcMap 10.2 was used to create a single LAS dataset file containing both LiDAR datasets, which was then filtered to include ground points only and create a 1-meter resolution DEM using the natural neighbor as the fill void method and the average as the interpolation method.

Table 1. LiDAR data acquisition parameters of both datasets collected over Robinson Forest.

| | Leaf-Off Dataset | Leaf-On Dataset |
|---|---|---|
| Date of Acquisition | April 23, 2013 | May 28- 30, 2013 |
| LiDAR System | Leica ALS60 | Leica ALS60 |
| Average Flight Elevation above Ground | 3,096 m | 196 m |
| Average Flight Speed | 105 knots | 105 knots |
| Pulse Repetition Rate | 200 KHz | 200 KHz |
| Field of View | 40º | 40º |
| Swath Width | 2,253.7 m | 142.7 m |
| Usable Center Portion of Swath | 90% | 95% |
| Swath Overlap | 50% | 50% |
| Average Footprint | 0.6 m | 0.15 m |
| Nominal Post Spacing | 0.8 m | 0.2 m |

## 3.3 Performance evaluation

To evaluate the performance of the tree segmentation approach, we compared the location of trees in the stem map created from field collected data with the location of LiDAR-derived tree locations. As stump locations seldom coincide with the location of the crown apexes (LiDAR-derived tree locations) due to leaning and irregular crown shape, the exact coordinates from the stem map were not used in the evaluation. Instead, we improved the tree detection evaluation procedure used by Kaartinen et al. (2012). A LiDAR-derived tree location matches with a stem map location if: i) the angle between the vertical projection of the 3D coordinates of the stump location and the 3D coordinates of the LiDAR-detected apex is within a given leaning threshold, and ii) the height difference is within a given threshold. If more than one LiDAR-derived tree location match with a stem map location or vice versa, only the best one is used.

A scoring system was developed to match multiple LiDAR-derived tree locations with the most appropriate stem map location. Three increasing leaning (5°, 10°, and 15°) and height difference (10%, 20%, and 30%) threshold levels with decreasing scores (100, 70, and 40) were considered (Table 2, Figure 5). A matrix with matching scores for all possible pairs of LiDAR-derived tree locations (rows) and stem map locations (columns) was then constructed. It was then processed by the Hungarian assignment algorithm (Kuhn, 1955) to produce the optimal matching assignment with the greatest overall matching scores.

Table 2. Leaning and height difference thresholds with associated scores considered for matching LiDAR-derived tree locations to stem map locations.

| Leaning threshold ($^0$) | Height difference threshold (%) | Score |
|---|---|---|
| 5 | 10 | 100 |
| 10 | 20 | 70 |
| 15 | 30 | 40 |
| > | > | 0 |

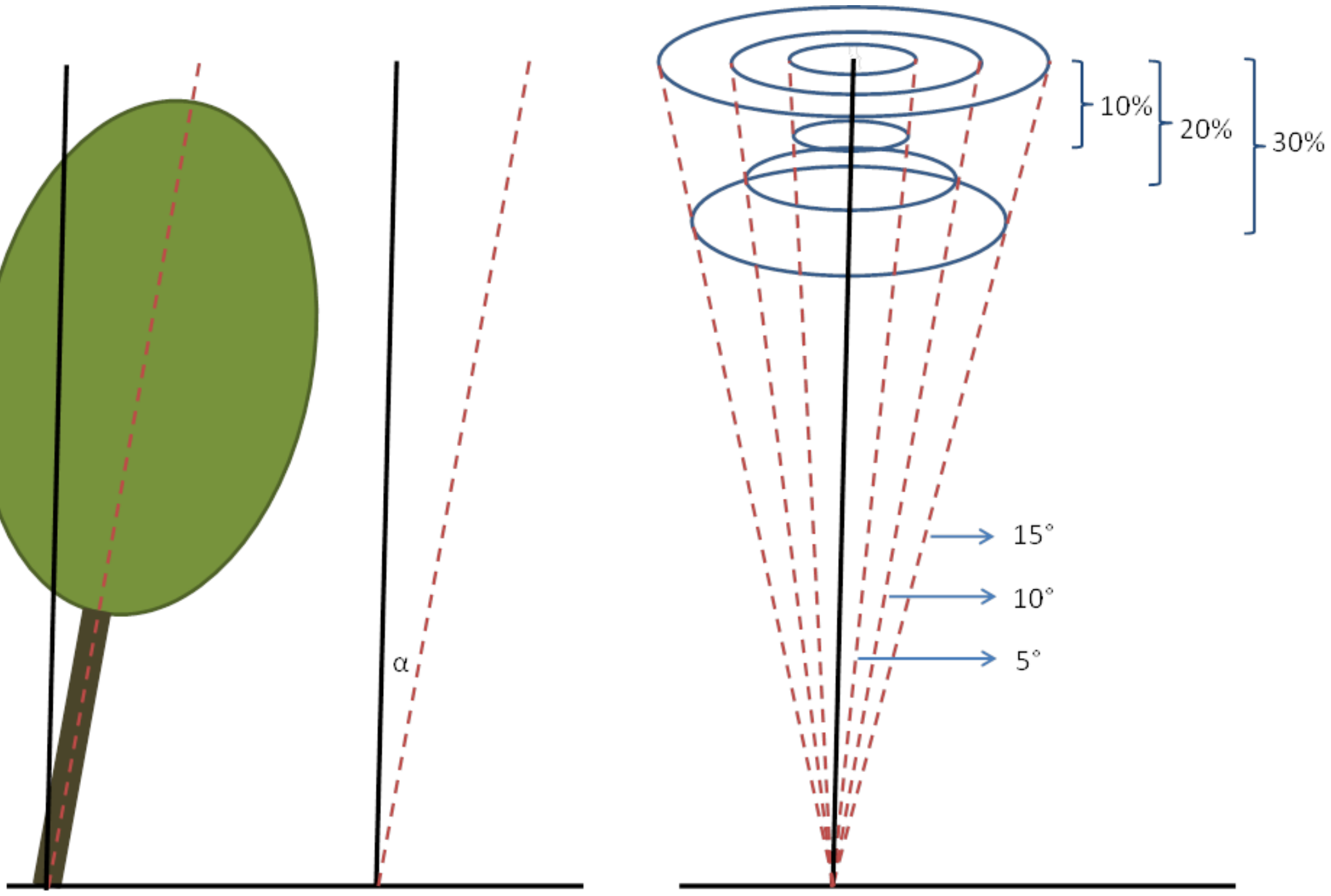

Figure 5. Calculation of leaning angle and distance difference used in the matching score system.

In the optimal assignment, a matched tree is an assigned pair of a LiDAR-derived tree location and a stem map location. An omission is a stem map location that remains unassigned (score=0). A commission is an unassigned LiDAR-derived tree location. The number of matched trees (MT) is an indication of the segmentation quality. The number of omission errors (OE) and commission errors (CE) indicate under- and over-segmentation, respectively. The accuracy of the approach was calculated in terms of recall (Re), precision (Pr) and F-score (F) using the following equations (Manning et al., 2008):

$$Re = \frac{MT}{MT + OE} \quad [6]$$

$$Pr = \frac{MT}{MT + CE} \quad [7]$$

$$F = 2 \times \frac{Re \times Pr}{Re + Pr} \quad [8]$$

Recall is a measure of the tree detection rate, precision is a measure of correctness of detected trees and the F-score indicates the overall accuracy taking omission and commission errors into account.

# 4 Results and Discussion

After visually inspecting LiDAR ground points and intensity values, boards (and thus plot centers) were clearly identified for only 23 permanent plots, which were considered for the evaluation of the tree-segmentation approach. Although the location of the remaining plots could be estimated by triangulation to clearly visible objects on the ground and the LiDAR data (e.g., large trees, rock formations, vegetation gaps, road features), they were not considered in the analysis to avoid mismatching exact plot locations and thus obscuring comparisons between the tree-segmentation approach and the field-collected data. Plots were located on all aspect orientations and on slopes ranging from 10% to 70%. An average of 13.2 trees were tallies per plot, with an average species diversity index (Shannon, 2001) of 1.47 (Table 3). The LiDAR point cloud over each plot included a 3.2-m (10 ft) buffer for capturing complete crowns of border trees.

Table 3. Summary plot level data collected from the 23 plots in the study area.

| Plot-Level Metric | | Min | Max | Average | Total | Percent of total |
|---|---|---|---|---|---|---|
| Slope | (%) | 10 | 70 | 41 | | |
| Aspect | $^{o}$ | 16 | 359 | 185 | | |
| Tree count | | 6 | 27 | 13.2 | 303 | |
| Dominant | | 0 | 3 | 0.6 | 14 | 4.6 |
| Co-dominant | | 0 | 10 | 3.4 | 78 | 25.7 |
| Intermediate | | 2 | 10 | 5.5 | 126 | 41.6 |
| Overtopped | | 0 | 15 | 3.1 | 72 | 23.8 |
| Dead | | 0 | 5 | 0.6 | 13 | 4.3 |
| Species count | | 3 | 9 | 5.6 | 33 | |
| Shannon diversity index | | 0.8 | 2.01 | 1.47 | | |
| Median tree height | (m) | 13.0 | 24.7 | 18.3 | | |
| Interquartile range of tree heights | (m) | 2.6 | 8.8 | 5.5 | | |

The accuracy of the tree-segmentation approach on trees in the 23 plots is presented in Table 4. On average, the tree detection rate of the segmentation approach was 72%, and 86% of detected trees were correctly detected. The overall accuracy in terms of the F-score was 77%. Recall values ranged from 31% to 100% and precision values ranged from 50% to 100%. In dense plots with a relatively large number of intermediate and overtopped trees, several trees were under-segmented resulting in relatively low recall values. For example, 6 of 19 and 0 of 11 intermediate and overtopped trees were detected in plots 4 and 11, respectively. However, all dominant and co-dominant trees in these two plots were detected. As expected, the three accuracy metrics were higher for dominant and co-dominant trees compared with intermediate and overtopped trees (Table 4). Recall increased to 94% for larger trees and decreased to 62% for smaller trees. Precision was more stable; it changed slightly about 1% from the overall 86%, 87% for larger trees and 85% for smaller trees. When considering all trees, the tree-segmentation approach was able to detect 100% of dominant, 92% of co-dominant, 74% of intermediate, and 38% of overtopped trees in the 23 plots. In addition, the approach was able to detect 39% of dead trees (Table 4).

Table 4. Summary of accuracy results of the tree segmentation approach on the 23 plots.

| | Number of Lidar detected / Field measured by tree class | | | | | Total number of matches and errors | | | Overall accuracy (%) | | | Accuracy by tree class group (%) | | | | | |
|---|---|---|---|---|---|---|---|---|---|---|---|---|---|---|---|---|---|
| | | | | | | | | | | | | D & C | | | I, O, & Dead | | |
| Plot | D[1] | C[2] | I[3] | O[4] | Dead | MT[5] | OE[6] | CE[7] | Re[8] | Pr[9] | F[10] | Re | Pr | F | Re | Pr | F |
| 1 | 0/0 | 3/3 | 6/10 | 1/3 | 0/0 | 10 | 6 | 3 | 62.5 | 76.9 | 69.0 | 100.0 | 75.0 | 85.7 | 53.8 | 77.8 | 63.6 |
| 2 | 1/1 | 3/3 | 4/4 | 2/6 | 0/0 | 10 | 4 | 1 | 71.4 | 90.9 | 80.0 | 100.0 | 100.0 | 100.0 | 60.0 | 85.7 | 70.6 |
| 3 | 0/0 | 3/3 | 4/4 | 0/5 | 0/1 | 7 | 6 | 1 | 53.8 | 87.5 | 66.6 | 100.0 | 75.0 | 85.7 | 40.0 | 100.0 | 57.1 |
| 4 | 1/1 | 4/4 | 3/4 | 3/15 | 2/3 | 13 | 14 | 0 | 48.1 | 100.0 | 65.0 | 100.0 | 100.0 | 100.0 | 36.4 | 100.0 | 53.3 |
| 5 | 1/1 | 4/4 | 9/9 | 6/7 | 0/0 | 20 | 1 | 0 | 95.2 | 100.0 | 97.5 | 100.0 | 100.0 | 100.0 | 93.8 | 100.0 | 96.8 |
| 6 | 2/2 | ½ | 3/3 | 0/0 | 0/0 | 6 | 1 | 1 | 85.7 | 85.7 | 85.7 | 75.0 | 100.0 | 85.7 | 100.0 | 75.0 | 85.7 |
| 7 | 0/0 | 9/10 | 2/8 | 0/3 | 0/0 | 11 | 10 | 4 | 52.4 | 73.3 | 61.1 | 90.0 | 75.0 | 81.8 | 18.2 | 66.7 | 28.6 |
| 8 | 1/1 | 5/6 | 5/8 | 0/1 | 0/0 | 11 | 5 | 1 | 68.8 | 91.7 | 78.6 | 85.7 | 85.7 | 85.7 | 55.6 | 100.0 | 71.4 |
| 9 | 0/0 | 2/2 | 7/9 | 2/3 | 0/0 | 11 | 3 | 1 | 78.6 | 91.7 | 84.6 | 100.0 | 66.7 | 80.0 | 75.0 | 100.0 | 85.7 |
| 10 | 0/0 | 1/1 | 2/2 | 3/6 | 1/5 | 7 | 7 | 0 | 50.0 | 100.0 | 66.7 | 100.0 | 100.0 | 100.0 | 46.2 | 100.0 | 63.2 |
| 11 | 1/1 | 4/4 | 0/8 | 0/3 | 0/0 | 5 | 11 | 2 | 31.3 | 71.4 | 43.5 | 100.0 | 71.4 | 83.3 | 00.0 | 00.0 | 00.0 |
| 12 | 0/0 | 4/4 | 3/4 | 2/3 | 0/0 | 9 | 2 | 4 | 81.8 | 69.2 | 75.0 | 100.0 | 57.1 | 72.7 | 71.4 | 83.3 | 76.9 |
| 13 | 0/0 | 3/3 | 7/7 | 0/0 | 0/0 | 10 | 0 | 1 | 100.0 | 90.9 | 95.2 | 100.0 | 100.0 | 100.0 | 100.0 | 87.5 | 93.3 |
| 14 | 0/0 | 2/2 | 3/3 | 1/1 | 0/0 | 6 | 0 | 0 | 100.0 | 100.0 | 100.0 | 100.0 | 100.0 | 100.0 | 100.0 | 100.0 | 100.0 |
| 15 | 0/0 | 9/9 | 3/3 | 0/0 | 0/1 | 12 | 1 | 2 | 92.3 | 85.7 | 88.9 | 100.0 | 81.8 | 90.0 | 75.0 | 100.0 | 85.7 |
| 16 | 1/1 | ½ | 5/8 | 3/6 | 0/0 | 10 | 7 | 0 | 58.8 | 100.0 | 74.1 | 66.7 | 100.0 | 80.0 | 57.1 | 100.0 | 72.7 |
| 17 | 0/0 | 4/4 | 6/6 | 2/2 | 1/1 | 13 | 0 | 4 | 100.0 | 76.5 | 86.7 | 100.0 | 66.7 | 80.0 | 100.0 | 81.8 | 90.0 |
| 18 | 3/3 | 0/0 | 1/3 | 0/0 | 0/0 | 4 | 2 | 1 | 66.7 | 80.0 | 72.7 | 100.0 | 100.0 | 100.0 | 33.3 | 50.0 | 40.0 |
| 19 | 2/2 | 0/2 | 2/4 | 0/1 | 0/0 | 4 | 5 | 4 | 44.4 | 50.0 | 47.0 | 50.0 | 66.7 | 57.1 | 40.0 | 40.0 | 40.0 |
| 20 | 0/0 | 2/2 | 4/6 | 0/0 | 0/0 | 6 | 2 | 2 | 75.0 | 75.0 | 75.0 | 100.0 | 100.0 | 100.0 | 66.7 | 66.7 | 66.7 |
| 21 | 0/0 | 2/2 | 4/5 | 2/4 | 1/1 | 9 | 3 | 0 | 75.0 | 100.0 | 85.7 | 100.0 | 100.0 | 100.0 | 70.0 | 100.0 | 82.4 |
| 22 | 1/1 | 1/1 | 6/6 | 0/1 | 0/1 | 8 | 2 | 0 | 80.0 | 100.0 | 88.9 | 100.0 | 100.0 | 100.0 | 75.0 | 100.0 | 85.7 |
| 23 | 0/0 | 5/5 | 2/2 | 0/2 | 0/0 | 7 | 2 | 3 | 77.8 | 70.0 | 73.7 | 100.0 | 83.3 | 90.9 | 50.0 | 50.0 | 50.0 |
| Average detection | 14/14 100% | 72/78 92.3% | 93/126 73.8% | 27/72 37.5% | 5/13 38.6% | 206/303 68.0% | 94/303 31.1% | 35/303 11.6% | 71.7 | 85.5 | 76.7 | 94.2 | 87.1 | 89.5 | 61.6 | 84.7 | 70.9 |

[1] Dominant, [2] Co-dominant, [3] Intermediate, [4] Overtopped
[5] Matched Trees, [6] Omission Errors, [7] Commission Errors, [8] Recall, [9] Precision, [10] F-score

As an example, Figure 6 shows the results of the tree segmentation performance for plot 8, 14, 15, and 22. Empty areas close to plot boundaries represent crowns of non-matched trees outside the plots (apex is outside of the boundary), which were removed from the analysis. Omissions in these empty areas (i.e., lower right side of plot 8) are intermediate and overtopped trees likely below dominant trees outside the plot boundary. As the LiDAR point clouds include buffer areas, several matched tree crowns extend beyond the plot boundary. Many crowns do not look circular because of the dense canopies and the fact that the crowns may be undercover to some extent. Two commissions can be observed in plot 15 where nine co-dominant trees are growing tightly in a small area.

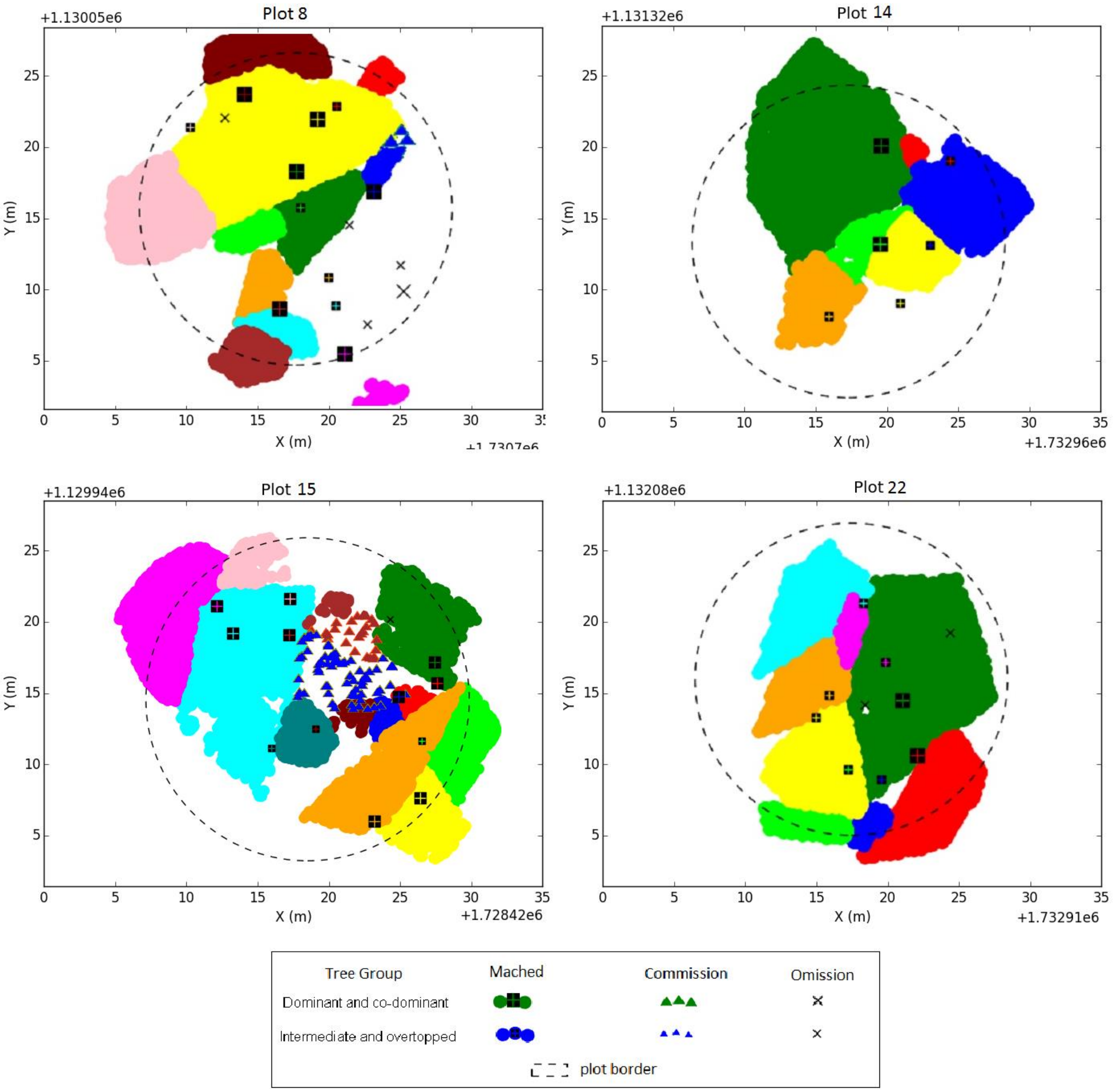

Figure 6. Aerial visualization of the tree segmentation results in four plots within the study area. Distinct colors represent matched tree crowns.

We evaluated relationships between accuracy metrics for each tree group (precision, recall, and F-score) and plot level attributes, i.e., average terrain slope, tree density, species diversity index, percentage of dominant, co-dominant, intermediate, overtopped, and dead trees, as well as median and interquartile range of tree heights (IQRH). None of the relationships for dominant and co-dominant group of trees was statistically significant. For the smaller group of trees, we observed negative correlations between recall and IQRH (P=0.004, $R^2$=0.33) and diversity index (P=0.03, $R^2$=0.2). Similarly, there was a negative correlation between F-score and IQRH (P=0.03, $R^2$=0.21). Also, a negative correlation between precision and percentage of dominant trees (P=0.02, $R^2$=0.24) was observed. These correlations indicate that in multi-story plots with large dominant trees, intermediate and suppressed trees are more difficult to detect. As the tree segmentation approach considers only LSPs, dominant and co-dominant trees can be easily detected. On the other hand, the crowns of intermediate and overtopped trees are only partially visible from above and in some cases completely underneath large tree crowns, making them harder to be detected. As future work, we consider improving the approach by creating vertical canopy layers through examination of the height distribution of LiDAR point as suggested by Duncanson et al. (2014) and Wang et al (2008). The correlations we observed between accuracy metrics and plot level attributes are weak and insignificant specially for larger group of trees, which likely indicates that the accuracy of the tree-segmentation approach is not sensitive to differences in stand and terrain structures of the study area. This demonstrates the robustness of the approach and increases its potential applications.

Other tree-segmentation studies in closed-canopy deciduous forests have reported tree detection accuracies of about 50% (Koch et al., 2006; Weinacker et al., 2004), 65% (Jing et al., 2012), and 72% (Hu et al., 2014) using similar evaluation metrics, which take both omissions and commissions into account. Vauhkonen et al. (2011) compared six different single tree detection methods on two deciduous forest sites. Performances were similar across sites; the average F-score of all methods was 57% where the maximum F-score was 64% and average recall and precision were 47% and 74%, respectively. Also, Duncanson et al. (2014) used a multilayered crown delineation approach, which correctly identified 70% of dominant trees, 58% of co-dominant trees, 35% of intermediate trees, and 21% of overtopped trees in a deciduous forest. Tree-segmentation accuracies from these previous studies in deciduous forests are slightly lower than the accuracy from our novel approach, which is an indicator of potential applicability of our study to deciduous forests with complex vegetation conditions.

## 5 Conclusions

Developing automated approaches to obtain tree-level information over large forested areas is increasingly important for accurate assessment, monitoring and management. Most of existing methods are forest type specific and applied to conifer forests. In this study, we developed a generalized tree-segmentation approach that uses small foot print LiDAR data and applied it to natural deciduous forests with complex structures. A significant advantage of our novel approach is that it does not require a priori knowledge of tree shapes and sizes. The approach retrieves local information, crown steepness and height of the vegetation, and uses it on-the-fly to enhance crown delineation.

Using an improved evaluation method, results showed that our approach was able to detect 72% of trees, and 86% of detected trees were correctly identified, resulting in an overall accuracy of 77%. Examining results by crown class, the approach detected 94% of dominant and co-dominant trees and 62% of intermediate and overtopped trees. Statistical analysis revealed similar accuracy levels across plots with different structures, which indicates the potential successful application of our approach to other forest types.

The main research challenge of the proposed tree segmentation approach was capturing heterogeneously shaped trees, and detecting intermediate and overtopped trees that may entirely be non-present within LSPs was not attempted explicitly. Future research will focus on incorporating multilayered segmentation method to improve detection of these smaller undercover trees. Our approach currently considers only LMs as the potential crown boundaries while other non-LM points might also represent boundaries, which in turn would result in omissions. Considering other points as potential crown boundaries (e.g., points showing distinct slope patterns on either side) can help to improve performance of the method. Although in general the number of omissions was larger than commissions, improving precision especially for larger trees can also enhance overall accuracy. A post-processing step to identify over-segmentation, using a scoring mechanism to distinguish branch foliage from an entire tree crown, as suggested in (Hu et al., 2014; Véga et al., 2014; Wolf and Heipke, 2007) and then merging over-segmented crown sections back to form entire tree crowns can help to reduce commissions.

## Acknowledgements

This work was supported by: 1) the Department of Forestry at the University of Kentucky and the McIntire-Stennis project KY009026 Accession 1001477, ii) the University of Kentucky Center for Computational Sciences, iii) the Kentucky Science and Engineering Foundation under the account number KSEF-3405-RDE-018, and iv) the National Science Foundation under Grant Number CCF-1215985.

Appendix A: Expected slope of a spherical surface

The slope of a spherical surface in degrees ranges between 0-90° (Figure A.1). Assuming the LiDAR surface points are uniformly distributed along the horizontal dimension, the expected value of the angle $\alpha = sin^{-1}x$ is calculated as follows.

$$\bar{\alpha} = \int_0^1 sin^{-1}x.dx = \frac{\pi}{2} - 1 = 32.7^{\circ} \qquad \textit{[A.1]}$$

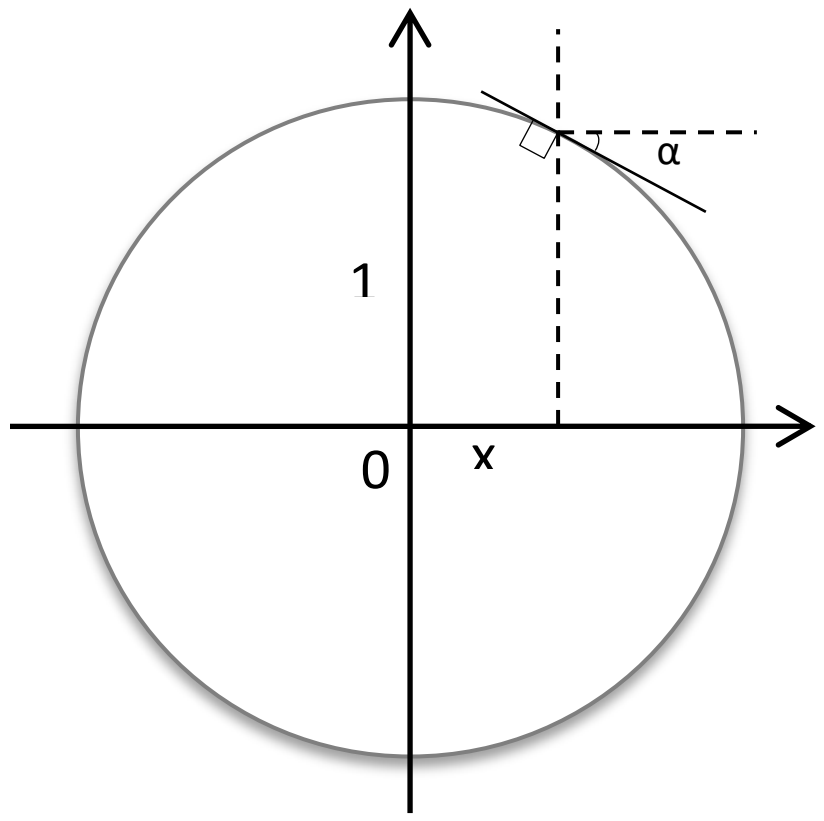

Figure A.1. The angle α representing the slope of the unit circle is a function of x.